\documentclass{ieeeaccess}
\IEEEoverridecommandlockouts

\usepackage{cite}
\usepackage{amsmath,amssymb,amsfonts}
\usepackage{graphicx}
\usepackage{textcomp}
\usepackage{booktabs}
\usepackage{array}
\usepackage{url}

\usepackage{bm}
\makeatletter
\AtBeginDocument{\DeclareMathVersion{bold}
\SetSymbolFont{operators}{bold}{T1}{times}{b}{n}
\SetSymbolFont{NewLetters}{bold}{T1}{times}{b}{it}
\SetMathAlphabet{\mathrm}{bold}{T1}{times}{b}{n}
\SetMathAlphabet{\mathit}{bold}{T1}{times}{b}{it}
\SetMathAlphabet{\mathbf}{bold}{T1}{times}{b}{n}
\SetMathAlphabet{\mathtt}{bold}{OT1}{pcr}{b}{n}
\SetSymbolFont{symbols}{bold}{OMS}{cmsy}{b}{n}
\renewcommand\boldmath{\@nomath\boldmath\mathversion{bold}}}
\makeatother

\newcommand{\sys}{\textsc{Scoped}}
\newcommand{\devser}{\textsc{Flat}}
\newcommand{\pixarm}{\textsc{pixel}}
\newcommand{\ci}[2]{[#1,\,#2]}

\begin{document}
\setlength{\textfloatsep}{12pt plus 2pt minus 4pt}
\setlength{\floatsep}{12pt plus 2pt minus 4pt}
\setlength{\intextsep}{12pt plus 2pt minus 4pt}
\history{This work has been submitted to the IEEE for possible publication. Copyright may be transferred without notice, after which this version may no longer be accessible.}
\doi{none (preprint)}

\title{Free the Language Model From the Vision Encoder: Semantic Serialization as a Perception Interface for Small Language Models}

\author{\uppercase{Cong Xu}\authorrefmark{1} AND \uppercase{Ravi Sankar}\authorrefmark{1}, \IEEEmembership{Life Senior Member, IEEE}}

\address[1]{iCONS Lab, Department of Electrical and Computer Engineering, University of South Florida, Tampa, FL 33620 USA}
\tfootnote{This work received no external funding. Question banks, per-item logs, perceived states, the frozen amendment ledger, and the analysis code are publicly released; see the Data and Code Availability statement.}

\markboth
{Xu and Sankar: Free the Language Model From the Vision Encoder}
{Xu and Sankar: Free the Language Model From the Vision Encoder}

\corresp{Corresponding author: Cong Xu (e-mail: cong@usf.edu).}

\begin{abstract}
End-to-end vision--language models (VLMs) bind visual competence to the scale of their language model: as the language model shrinks, perception and reasoning degrade together. We study an embodied scene question-answering (QA) interface in which vision never enters the language model. A frozen perception stack detects and ranges objects; a deterministic semantic serializer compiles the perceived state, errors included, into decision-aligned text; an unmodified text-only large language model (LLM) answers. On a visible-scope-matched, occlusion-audited campus-robot benchmark, under a prospectively frozen criterion, the serialized interface, using detectors fine-tuned in-domain within each fold, outperforms a zero-shot VLM whose language model has the same 7B scale ($0.7892$ vs $0.7462$), with a larger margin at 3B ($0.7673$ vs $0.6913$). Preregistered decoupling experiments show the gain survives paraphrase, attributing it to decision-aligned computation rather than answer-string leakage, while novel judgment vocabularies bound its scope. The advantage grows as the reader shrinks to 1.5B and reverses at 0.5B, and a ground-truth oracle locates the reader-capability floor. Under matched task supervision the interfaces converge: a VLM fine-tuned with low-rank adaptation (LoRA) overtakes the zero-shot system but only ties an equally supervised text reader ($0.8441$ vs $0.8396$, no statistically resolved difference), and both routes remain perception-bound. Reported perception parameters are comparable to those of the VLM's vision tower, and total compute is not smaller.
\end{abstract}


\titlepgskip=-21pt

\maketitle

\section{Introduction}
\label{sec:intro}

\PARstart{M}{ultimodal} large language models (LLMs) answer questions about images by prepending visual tokens, the output of a Vision Transformer (ViT) image encoder \cite{dosovitskiy2021vit}, to the language model's input \cite{li2023blip2,liu2023llava,bai2025qwen25vl}. This coupling has a consequence for small models: as the decoder (the language model that generates the answer) shrinks, visual grounding and language reasoning degrade together, because both live in the same weights. Substantial visual-token redundancy has been reported in several large vision--language models (LVLMs) \cite{chen2024fastv}, and benchmarks find deficits in exactly the skills embodied platforms need: counting \cite{paiss2023countbench}, fine perception \cite{fu2024blink}, and spatial grounding \cite{tong2024eyeswide,liu2023vsr,kamath2023whatsup}. The coupling costs most precisely on the embodied platforms that most need a small model.

We study the opposite factorization: vision stays entirely outside the language model. Frozen vision modules perceive; a deterministic \emph{semantic serializer} compiles their output into a compact, physically grounded textual state; an off-the-shelf text-only LLM, which we call the \emph{reader}, reads it and answers, never seeing a pixel or a visual token. The reader, the component one would distill or shrink, spends none of its capacity on visual grounding. All abbreviations, terminology, and numerical conventions used in this paper are provided in Section~\ref{sec:terms}.

Prior ``looking''-vs-``reading'' comparisons were confounded twice over: text descriptions usually encode a privileged subset of the scene, and serialized vocabularies can leak answer strings. We address the first confound with scope matching and the second with preregistered decoupling probes (Section~\ref{sec:mechanism}). Concretely, we build a \emph{visible-scope-matched} benchmark on the UT Campus Object Dataset (CODa)~\cite{zhang2024coda}: five-family questions generated only from the camera-visible ground-truth set defined in Section~\ref{sec:bench}, \eqref{eq:vgt}. The pixel arm receives the raw image $I_t$; our \emph{real system} receives $\sigma(\hat V_t)$ with $\hat V_t = P(I_t)$, misses, false positives, and range error included; and an \emph{oracle} arm receives $\sigma(V^{GT}_t)$. Scope is matched; representational maturity is not, so we keep system-level and interface-level conclusions apart throughout the paper. One restriction up front: the interface is task-aware, assuming decision vocabularies known at deployment (typical of fielded systems); Section~\ref{sec:mechanism} measures the cost of leaving that regime.

\textbf{Contributions.}
\begin{enumerate}
\item A visible-scope-matched and occlusion-audited question-answering (QA) benchmark on CODa with oracle and real-perception arms, two frame-disjoint banks (1366 development, 1328 confirmatory), and frozen paraphrase (N1) and novel-judgment (N2) probe sets.
\item A system-level, criterion-frozen zero-shot comparison: the serialized interface exceeds a vision--language model (VLM) of the same language-model size by $+0.0429$ accuracy at 7B (95\% confidence interval (CI) $\ci{+0.0073}{+0.0690}$) and by $+0.0761$ at 3B ($\ci{+0.0403}{+0.0996}$), replicated on a current model family at the 4B size.
\item A mechanism account: the gain survives paraphrase, is dominated by boundary-aligned precomputation, and is bounded by novel judgment vocabularies.
\item A reader-scale study locating both the advantage (down to a 1.5B reader) and its floor (0.5B), with a ground-truth oracle separating perception-bound from reader-bound regimes.
\item A supervision study with a full supervision ledger: a VLM tuned with low-rank adaptation (LoRA) overtakes the zero-shot system, but an equally supervised text reader statistically ties it ($-0.0045$, $\ci{-0.0368}{+0.0177}$), so under matched supervision the interface difference dissolves rather than reverses.
\end{enumerate}

\section{Abbreviations, Terminology, and Numerical Conventions}
\label{sec:terms}

This paper sits between robotics perception and language-model research, and its evidence is statistical. So that a reader from either side can follow every claim, Table~\ref{tab:abbr} lists the abbreviations used in the paper, Table~\ref{tab:terms} defines the technical terms in plain language, and this section outlines the numerical conventions.

\begin{table}[t]
\centering
\caption{Abbreviations used in this paper.}
\label{tab:abbr}
\setlength{\tabcolsep}{4pt}
\scriptsize
\begin{tabular}{>{\raggedright\arraybackslash}p{1.55cm}p{6.25cm}}
\toprule
abbreviation & meaning \\
\midrule
0.5B, 1.5B, 3B, 4B, 7B, 8B & model size in billions of trainable parameters (for text readers: the language model; for VLMs: the whole model) \\
$^{*}$ & a difference whose 95\% confidence interval excludes zero (``resolved'') \\
cf / dev & confirmatory bank (1328 questions, used once, for the final test) / development bank (1366 questions, used to build the system) \\
CI & confidence interval (always 95\%, cluster bootstrap) \\
CODa & UT Campus Object Dataset \cite{zhang2024coda} \\
D0--D3 & serializer variants: D0 = \sys{} (shipped); D1 = numeric ranges; D2 = numeric plus alternative bands; D3 = alternative bands only \\
\devser{} / \sys{} & the preregistered serializer (one stream for every question family) / the shipped serializer (stream scoped to the family) \\
FOV & field of view of the front camera \\
FT, FT-VLM, FT-Text, FT-3B, FT-7B & fine-tuned arms: the VLM fine-tuned on task data (FT-VLM; FT-3B and FT-7B for its two sizes) and the text reader fine-tuned on the same data (FT-Text) \\
GT & ground truth: the dataset's human annotations \\
LiDAR & laser range sensor; used here only for the occlusion audit, never as system input \\
LLM / VLM / LVLM / MLLM & large language model / vision--language model (an LLM that also takes images; ``large'' and ``multimodal'' variants of the name are synonyms here) \\
LoRA / QLoRA & low-rank adaptation: fine-tuning by training a small set of added weights; QLoRA is its memory-saving quantized variant \\
N1 / N2 & probe banks: N1 = every confirmatory question reworded; N2 = new question types outside the serializer's vocabulary \\
OOF & out-of-fold: evaluated by a model that never saw the question's recording \\
P68 / P95 & 68th / 95th percentile \\
\pixarm{} & the baseline arm: a VLM receiving the raw image \\
pp & percentage points \\
QA & question answering \\
ViT & Vision Transformer, the image encoder inside a VLM \\
VLA & vision--language--action model (a VLM that outputs robot actions) \\
YOLO11 & a real-time object detector \cite{jocher2024yolo11} \\
\bottomrule
\end{tabular}
\end{table}

\begin{table}[t]
\centering
\caption{Technical terms, in plain language, as used in this paper.}
\label{tab:terms}
\setlength{\tabcolsep}{4pt}
\scriptsize
\begin{tabular}{>{\raggedright\arraybackslash}p{1.9cm}p{5.9cm}}
\toprule
term & meaning \\
\midrule
language model, decoder & a neural network that produces text one token at a time; the ``decoder'' is the part that generates the answer. Our reader is a text-only decoder \\
token, visual token & the unit a language model reads; text is split into word pieces, and a VLM converts image patches into ``visual tokens'' that occupy the same input positions \\
vision encoder & the network (a ViT) that turns an image into visual tokens; in a VLM it is wired to the language model \\
end-to-end VLM & a single network that takes pixels and text directly and answers; our baseline (the \pixarm{} arm) \\
reader & the text-only language model that reads the serialized state and answers; it never sees the image \\
serializer, serialized (compiled) state & a deterministic program, not a model, that turns the detected objects (class, bearing, range, confidence) into a short text; ``compiled'' stresses that closed-form quantities are computed before the reader sees them \\
question family & one of five question types: counting, direction, nearest-class, nearest-distance, path-object \\
decision vocabulary, band, sector & the fixed answer words of a family: distance bands (``under 5\,m'', ``5 to 15\,m'', ``over 15\,m'') and sectors (left, center, right) \\
bank & a fixed list of questions; see cf / dev in Table~\ref{tab:abbr} \\
probe bank & a bank built to test a specific threat: N1 (same questions, reworded) tests wording dependence; N2 (new judgment types) tests leaving the known vocabulary \\
arm & one evaluated configuration (model + input form), e.g., \pixarm{}, \sys{}, GT oracle \\
oracle & an arm that receives ground-truth object states instead of perceived ones; an upper bound on what the interface can deliver \\
blind arm & the reader given the question and options but no scene at all; measures what wording alone gives away \\
zero-shot & the model is used exactly as released, with no training on our data \\
fine-tuning, gradient samples & updating (a small set of) weights on task examples; ``16k gradient samples'' means 16 000 training examples \\
prompt & the text handed to the model: serialized state, question, options, and a fixed answer-format instruction \\
greedy decoding ($T{=}0$) & the model always emits its single most likely next token; outputs are deterministic \\
parse rate, unparsed & fraction of outputs containing a readable answer letter; an unparsed output is scored as wrong \\
accuracy & fraction of questions answered correctly, on a 0--1 scale; chance is $1/3$ with three options \\
paired difference $\hat\Delta$ & accuracy of one arm minus another on the same questions \\
cluster bootstrap, resolved / unresolved & the uncertainty of $\hat\Delta$ is estimated by resampling the 21 recording sequences 20\,000 times; a difference is ``resolved'' ($^{*}$) if its 95\% CI excludes zero, ``unresolved'' otherwise \\
frozen criterion, preregistration, amendment & the decision rule and analysis plan were written and hash-committed before the data were seen; later additions (A3--A5) were frozen the same way \\
sequence-rotating $K$-fold, out-of-fold & detectors are trained on some recordings and tested on others, rotated so that no question is ever answered by a model that saw its own recording \\
perception-bound / reader-bound & accuracy is limited by what the detectors deliver / by what the language model can do even with a perfect state \\
percentile (P68, P95) & the value below which 68\% / 95\% of measurements fall \\
attenuation & $1-\hat\Delta_{\mathrm{N1}}/\hat\Delta_{\mathrm{cf}}$: the fraction of the advantage lost when questions are reworded \\
supervision ledger & a table stating how many training examples every arm received, so that comparisons are never between unequally trained systems \\
\bottomrule
\end{tabular}
\end{table}

\textbf{Numerical conventions.} Every accuracy is a fraction of questions answered correctly on the named bank (so $0.7892$ on the confirmatory bank means 1048 of 1328 questions). $\hat\Delta$ is a paired difference between two arms on the same questions, with a 95\% cluster-bootstrap CI; $^{*}$ marks a CI that excludes zero. Confirmatory-bank accuracies and contrasts are quoted to four decimals, exactly as tabulated; development-bank (exploratory) values are quoted to three decimals, the precision at which they were logged. Model sizes are given in billions of parameters. Differences stated in percentage points (pp) are $100\times$ the corresponding accuracy differences.

\section{Related Work}
\label{sec:related}

\subsection{How Mainstream MLLMs Encode Vision}
Virtually every production multimodal LLM couples a contrastively pre-trained ViT \cite{dosovitskiy2021vit,radford2021clip,zhai2023siglip} to a language model through a learned multilayer perceptron (MLP) bridge, whether cross-attention \cite{alayrac2022flamingo}, a resampler \cite{bai2023qwenvl,dai2023instructblip}, a visual-expert module \cite{wang2023cogvlm}, or an MLP projection \cite{li2023blip2,liu2023llava,liu2024llava15,wang2024qwen2vl,bai2025qwen25vl,chen2024internvl}, with connector and recipe design studied systematically \cite{cha2024honeybee,mckinzie2024mm1,karamcheti2024prismatic,tong2024cambrian}. A smaller family drops the encoder but keeps a learned visual pathway \cite{bavishi2023fuyu,diao2024eve,luo2024monointernvl} or fuses discrete visual tokens early \cite{chameleon2024}, and most bridge tokens are redundant at inference \cite{chen2024fastv}. All of these interfaces are \emph{learned and continuous}: the LLM must decode perception from embeddings. Ours is the opposite limit: deterministic, symbolic, auditable text into an unmodified LLM.

\subsection{Neuro-Symbolic and Programmatic Vision}
Recovering structured scene representations and executing symbolic programs go back to Neural-Symbolic VQA \cite{yi2018nsvqa}; VisProg \cite{gupta2023visprog} and ViperGPT \cite{suris2023vipergpt} compose deterministic visual modules with an LLM; Set-of-Mark \cite{yang2023som} writes symbolic anchors into the image; BLINDER \cite{nottingham2024blinder} learns to select concise state descriptions for LLM actors. We differ in three ways: compilation is deterministic code, not generated programs; fields are aligned to the decision vocabulary; and the comparison runs under matched scope, real perception errors, and a frozen criterion.

\subsection{VLMs for Embodied Platforms; Scene-as-Text}
PaLM-E and RT-2 bring encoder-based VLMs to robot decision-making \cite{driess2023palme,brohan2023rt2}, with open VLA successors \cite{kim2024openvla}; DriveVLM does so for driving \cite{tian2024drivevlm}; SpatialVLM adds spatially supervised pre-training while keeping the encoder \cite{chen2024spatialvlm}. A parallel line serializes object states into LLM prompts: GPT-Driver \cite{mao2023gptdriver}, object-level vector states fused into the decoder \cite{chen2024drivingllm}, interpretable end-to-end driving \cite{xu2024drivegpt4}, closed-loop language-conditioned driving \cite{shao2024lmdrive}, and 3D scene-graph planners \cite{rana2023sayplan,gu2024conceptgraphs}. These systems establish feasibility, but are evaluated against weak or information-mismatched baselines, leaving open whether text wins by format or by privileged content. Embodied QA benchmarks with scope discipline include nuScenes-QA \cite{qian2024nuscenesqa} (built on nuScenes \cite{caesar2020nuscenes}), DriveLM \cite{sima2024drivelm}, and Talk2BEV \cite{choudhary2024talk2bev}. We contribute the interface-level comparison these lines leave open, extended across question wording, reader scale, and supervision, against prior- and current-generation baselines \cite{qwen3vl2025}.

\section{Visible-Scope-Matched Benchmark}
\label{sec:bench}

\subsection{Data and Question Generation}
CODa~\cite{zhang2024coda} provides campus-robot front-camera images with ego-frame 3D boxes: an embodied, non-automotive setting. Across 21 recording sequences we generate one multiple-choice question per frame from five families (3 options; chance $=1/3$): \emph{counting}, \emph{direction}, \emph{nearest-class}, \emph{nearest-distance}, and \emph{path-object}. Questions are asked only about objects in the camera's field of view within 40\,m,
\begin{equation}
V^{GT}_t=\{e_i: \mathrm{inFOV}(e_i),\; r_i\le 40\,\mathrm{m}\},
\label{eq:vgt}
\end{equation}
so the image arm is never asked about what it cannot see. GT-derived controls serialize exactly the set in \eqref{eq:vgt}; the real system serializes its \emph{own} perceived set $\hat V_t$, never the full $360^\circ$ annotation.

The released bank retains a fourth abstention option, which is never the gold answer in either bank; the reported forced-choice track removes it and reletters the remaining three options, so chance is $1/3$. Throughout, the nearest-distance family is \emph{distance-band classification} over three fixed bands; we do not claim general metric estimation.

\subsection{Audit Gates}
\label{sec:audits}
A surface-cue audit gate checks option-length, gold-position, and length biases before either bank is frozen; a \emph{blind} arm, which sees the question and options but no scene, bounds the residual prior (the accuracy obtainable from wording alone) at $0.4142$ (3B reader) and $0.4089$ (7B reader) on the confirmatory bank, only modestly above the $1/3$ chance level (per-family breakdowns in Appendix~\ref{app:perfam}).

\textbf{Occlusion audit.} In-FOV membership does not guarantee image visibility, so we audit with the dataset's LiDAR: requiring ${\ge}10$ points inside the 3D box and ${\ge}20$\,px projected height affects $2.3\%$ of question-relevant objects, and recomputing the primary contrasts on the occlusion-clean subset ($n{=}1298$) leaves them essentially unchanged ($+0.0778^{*}$/$+0.0385^{*}$). Full-bank numbers are primary; the audit is in the supplement.

\subsection{Frozen Probe Sets}
Two probe banks were frozen, hashes committed, before any arm ran on them. \emph{N1} rewords every confirmatory stem and option through a deterministic paraphrase table, leaving boundaries and gold answers unchanged: it tests whether any advantage depends on shared surface wording. \emph{N2} poses 3974 questions whose judgment logic and boundaries (7/22\,m bands, cross-class ordinal comparisons, count-within-radius) appear in no serializer vocabulary: it measures the cost of leaving the decision vocabulary the serializer was built for.

\subsection{Arms}
\pixarm{}: Qwen2.5-VL \cite{bai2025qwen25vl} receives the raw image plus the question. \emph{Typed}: a text-only Qwen2.5 \cite{yang2024qwen25} receives the serialized state plus the question. Both run at the 7B and 3B sizes. A development-bank \emph{GT-typed} oracle serializes ground-truth boxes ($0.955$); caption and blind controls complete the ladder.

Even with ground-truth perception the serialization matters: at fixed GT content \devser{} beats an older serializer and a prose caption ($0.955$/$0.857$/$0.804$; development bank, exploratory; supplement); the oracle's 21\,pp margin over the pixel arm is what real perception must earn.

\section{Perception-to-Text System}
\label{sec:system}

\subsection{Perception Stack (Front Stereo Pair; Frozen at Inference)}
\label{sec:perception}
Two YOLO11 detectors \cite{jocher2024yolo11}, fine-tuned per fold on training sequences only (Section~\ref{sec:protocol}), are ensembled; all vision components are frozen at inference time. A candidate detection $d$ is accepted if either detector is confident alone or both agree weakly:
\begin{equation}
\max_i c_i(d) \geq \tau_h{=}0.50
\quad\text{or}\quad
\min_i c_i(d) \geq \tau_\ell{=}0.35 .
\label{eq:accept}
\end{equation}
Slicing-aided tiling \cite{akyon2022sahi} recovers small objects (border detections dropped). An open-vocabulary detector \cite{fu2025llmdet}, from the grounded-detection line \cite{liu2024groundingdino}, is \emph{demoted} to absence recovery (at most two objects of a class, only if the ensemble found none, confidence $\geq 0.60$, always tagged low-confidence) because in smoke tests direct fusion flooded the state with phantom objects, up to 86 detections of a single class (Bench) in a frame whose ground truth contained none.

For each accepted object with bounding-box bottom row $v_b$, pixel height $h_{\mathrm{px}}$, and horizontal center $u$, we compute four range estimates: ground-plane geometry $r_{\mathrm{grd}} = f_y h_{\mathrm{cam}} / (v_b - c_y)$ from the calibrated camera height; stereo disparity $r_{\mathrm{stt}}$; a class-height prior $r_{\mathrm{hgt}} = f_y H_k / h_{\mathrm{px}}$; and monocular metric depth $r_{\mathrm{uni}}$ \cite{piccinelli2024unidepth}. The fused range and bearing are
\begin{equation}
\begin{split}
\hat{r} &= \mathrm{med}\{r_{\mathrm{grd}},\, r_{\mathrm{stt}},\, r_{\mathrm{hgt}},\, r_{\mathrm{uni}}\},\\
\hat\theta &= \arctan\!\big((c_x - u)/f_x\big),
\end{split}
\label{eq:fuse}
\end{equation}
the median limiting the influence of any single-cue failure, though a single gross outlier still shifts the mid-pair mean. The bearing is a purely physical mapping through the intrinsics; nothing is learned. Absence-recovery proposals must also satisfy $r_{\mathrm{grd}} \in [\tfrac12 r_{\mathrm{uni}}, 2 r_{\mathrm{uni}}]$, a plausibility check between the two \emph{height-independent} cues. Deduplication merges detections $d_i,d_j$ if and only if
\begin{equation}
k_i{=}k_j \;\wedge\; |\hat\theta_i{-}\hat\theta_j| < 2^\circ \;\wedge\; \max(\hat r_i,\hat r_j)/\min(\hat r_i,\hat r_j) < 1.3,
\label{eq:dedup}
\end{equation}
and objects beyond $42$\,m are dropped. Applying \eqref{eq:accept}--\eqref{eq:dedup}, we obtain a perceived state $\hat{\mathcal{S}} = \{(k_i, \hat\theta_i, \hat r_i, \gamma_i)\}$ of class, bearing, range, and confidence tag $\gamma_i \in \{\textsf{conf}, \textsf{unc}\}$. \emph{No component is retrained between system versions.}

\subsection{Semantic Serializer: Compute What Can Be Computed}
\label{sec:serializer}
The serializer is deterministic code, not a model: $\sigma : (\hat{\mathcal{S}}, \text{question family}) \to$ text. Its principle came from the error attribution of Section~\ref{sec:attribution}: the reader is \emph{softly robust} to perception noise yet \emph{clumsy at arithmetic over clean numbers}. So the serializer computes every closed-form field in the decision's vocabulary, applied to the fused range and bearing of \eqref{eq:fuse}: a band map $\beta(r)$, ``under 5\,m'' / ``5 to 15\,m'' / ``over 15\,m'' at thresholds 5 and 15\,m, and a side map $\psi(\theta)$, ``left'' / ``center'' / ``right'' at $\pm 15^\circ$, with $\theta$ positive counterclockwise ($\theta>0$ left of the optical axis), matching the generator's ego-frame convention.

Band strings and thresholds are identical by construction to the generator's. Section~\ref{sec:mechanism} decouples this identity with serializer variants D1 (numeric ranges, selections retained), D2 (numeric plus alternative-vocabulary bands at 4/12\,m), and D3 (alternative bands only). The sector rule is \emph{not} shared (generator: image column; serializer: bearing); the four distinct agreement quantities this induces, object-level ($0.986$), selected-instance, and question-level under each rule ($0.553$/$0.761$), are separated in the supplement, the gap being instance selection and missed classes, not bearing error. We report the configuration as run, with the generator-rule variant as a labeled secondary analysis.

Three field types make up the stream. First, a ``nearest of type'' line per class with $\beta(\hat r)$ pre-banded into the exact option wording. Second, counts split by confidence tag, so the reader can weigh absence recovery. Third, borderline flags within $\varepsilon$ of a boundary, with $\varepsilon$ derived from the measured per-band range-error distribution (its 68th percentile, P68) so the flag width tracks the instrument's actual uncertainty. Everything ambiguous stays in natural language for the reader to aggregate: \emph{compile every closed-form quantity, leave the reader only what is genuinely ambiguous}. We write \sys{} for the shipped serializer and \devser{} for the preregistered one it replaced (\devser{}: one stream for every family; \sys{}: scoped to the family).

\emph{A verdict needs its appeal.} A closed-form selection over the \emph{perceived} state is correct on only $0.61$ of items and, when wrong, usually names a class outside the three options ($0.96$), so \sys{} ships every selection with its runner-up (the gold lies in the compiler's top two on $0.82$ of items). Compute what can be computed, and emit what the reader needs to determine when the computation was wrong.

\emph{Scope the stream to the decision.} $\sigma$ may read the question \emph{family}, known at deployment, but never the answer or the option set. Only counting needs the per-object rows: scoping them to that family cuts the median stream from $1842$ characters to about $720$ for the four families that do not need the rows (counting keeps them, about $1860$) and, with the selections above, is what lets the advantage survive a 3B reader (Section~\ref{sec:results}).

\subsection{Reader}
The reader is an unmodified, frozen Qwen2.5-7B text model \cite{yang2024qwen25} with a fixed answer-format prompt: no fine-tuning, few-shot examples, or logit surgery. A different-family reader of the same scale is evaluated in Section~\ref{sec:compute}.

\section{Prospectively Frozen Protocol}
\label{sec:protocol}

\textbf{Frozen criterion.} Before any confirmatory run the decision criterion was frozen. With correctness indicators $z_q^{\mathrm{sys}}, z_q^{\mathrm{pix}}$ grouped into the 21 sequence clusters, we resample clusters with replacement ($B{=}20000$, fixed seed) and compute
\begin{equation}
\hat\Delta = \tfrac{1}{N}\textstyle\sum_{q} \big(z_q^{\mathrm{sys}} - z_q^{\mathrm{pix}}\big),
\label{eq:boot}
\end{equation}
declaring a win if and only if $\mathrm{CI}^{95\%}_{\mathrm{lo}}(\hat\Delta) > 0$ over the bank's $N$ questions. Clustering respects near-duplication within a recording; pairing removes between-scene variance. Throughout the paper, $^{*}$ marks a 95\% CI excluding zero.

\textbf{Two co-primary endpoints.} 7B and 3B were both pre-specified as primary before the confirmatory bank existed; the headline claim is the conjunction, an intersection--union test needing no multiplicity adjustment ($\max p = 0.025$ over the same resampling; $p$ values by CI inversion; see the supplement).

\textbf{Development and confirmatory banks.} The \emph{development} bank (1366 questions) is where the stack was built and seven serializer versions were compared; everything measured there is exploratory. The shipped serializer is confirmed on a bank generated afterwards by the same code: 1328 questions, one per frame, each at least $1.5$\,s from every development frame (median separation $2.1$\,s, 95th percentile $4.7$\,s). It is \emph{temporally} disjoint, not sequence-disjoint: held-out time within the same 21 recordings, not external validation. A scene-content audit bounds near-duplication (no shared object sets; median displacement $2.32$\,m; $0.030$ within $1$\,m); the full composition and separation audit is in Appendix~\ref{app:bank}. No version, threshold, or prompt was touched after generation.

\textbf{Leak-free K-fold.} Since frames within a sequence are near-duplicates, we use $K{=}4$ \emph{sequence-rotating} folds: detectors and calibrations are fit per fold on training sequences only, and every question is answered by the fold model that never saw its sequence. All system numbers are out-of-fold.

\textbf{Amendment discipline.} Following prospectively frozen analysis plans \cite{nosek2018prereg} and reproducibility reporting practice \cite{pineau2021reproducibility}, amendments were frozen documents written and hash-committed before the runs they governed. Amendments A3 to A5 added the decoupling variants and probe banks, the reader-scale sweep, and the supervised arms with every decoding deviation. The full ledger, with commit hashes, per-amendment predictions, and the cases where the discipline caught a spurious out-of-fold advantage, is in the supplement.

\textbf{Instrument validation.} Accuracy is confounded with format compliance, so every arm reports its parse rate, and failures are plotted in run order. Table~\ref{tab:instr} gives, for each of the four primary arms, the accuracy, the accuracy restricted to parsed outputs, the parse rate, and the longest contiguous run of unparsed outputs; the last column is the diagnostic: a long contiguous unparsed block indicates a serving-state fault, not format non-compliance. Two such faults were caught and repaired by a reload-on-degenerate-output instrument; the broken 3B pixel arm would otherwise have reported $+0.601$, $8\times$ the true effect. Decoding is greedy ($T{=}0$); five repeated runs bound same-machine variation at ${\pm}0.0015$, 29--51$\times$ below the reported contrasts, and every paired contrast stays inside one machine. The full narrative and plots are in the supplement.

\begin{table}[t]
\centering
\caption{Instrument health, confirmatory bank (1328 items per arm). Unparsed outputs count as incorrect. The two same-decoder VL-on-text auxiliary arms (parse $0.974$/$0.968$) are read on parsed items ($+0.058$/$+0.120$, same sign; supplement). $^{\dagger}$After the reload re-run.}
\label{tab:instr}
\setlength{\tabcolsep}{4pt}
\footnotesize
\begin{tabular}{lcccc}
\toprule
arm & acc & acc$\mid$parsed & parse & longest unparsed run \\
\midrule
\sys{} 7B & 0.7892 & 0.7951 & 0.9925 & 1 \\
\pixarm{} 7B & 0.7462 & 0.7462 & 1.000 & 0 \\
\sys{} 3B & 0.7673 & 0.7714 & 0.9947 & 1 \\
\pixarm{} 3B$^{\dagger}$ & 0.6913 & 0.6928 & 0.9977 & 1 \\
\bottomrule
\end{tabular}
\end{table}

\section{Results}
\label{sec:results}

\begin{figure}[t]
\centering
\includegraphics[width=0.90\linewidth]{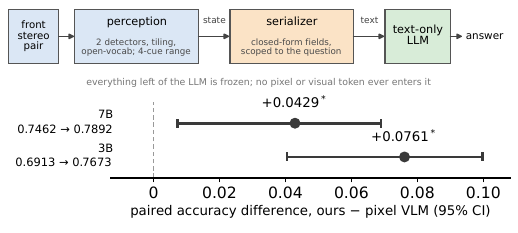}
\caption{\emph{Top:} the stack. Frozen vision modules perceive from the front stereo pair; a deterministic serializer compiles the perceived state into decision-aligned text; an unmodified text-only LLM answers; the baseline replaces everything left of the LLM with a ViT encoder over the monocular keyframe. \emph{Bottom:} confirmatory-bank paired accuracy differences with cluster-bootstrap 95\% CIs (out-of-fold; row labels give pixel $\to$ ours accuracies; parse rates in Table~\ref{tab:instr}).}
\label{fig:main}
\end{figure}

\subsection{Main Result: The Zero-Shot Regime}
Fig.~\ref{fig:main} summarizes the study: the top panel shows the system under test (frozen perception, deterministic serializer, text-only reader) against the baseline that replaces everything before the language model with a vision encoder; the bottom panel plots the two primary contrasts with their confidence intervals. On the 1328 confirmatory questions, one per frame, over frames no version was ever measured on, \sys{} scores $\mathbf{0.7892}$ versus $\mathbf{0.7462}$ for \pixarm{} at 7B ($\Delta = +0.0429$, 95\% CI $\ci{+0.0073}{+0.0690}$) and $\mathbf{0.7673}$ versus $\mathbf{0.6913}$ at 3B ($+0.0761$, $\ci{+0.0403}{+0.0996}$), meeting the criterion of \eqref{eq:boot}, frozen before the bank existed (Fig.~\ref{fig:main}). Per family: nearest-distance $+0.2362^{*}$ (3B) and $+0.1890^{*}$ (7B), counting $+0.1358^{*}$ at both sizes, nearest-class a resolved loss at 7B ($-0.0617^{*}$), direction and path-object unresolved (full per-family contrasts with CIs in Appendix~\ref{app:perfam}). The claim is therefore narrower than ``wins everywhere'': the gain concentrates in the two families that reduce to deterministic arithmetic over the state.

The result replicates on a current-generation family under identical decoding: Qwen3-4B reading the serialized state beats Qwen3-VL-4B by $+0.0595^{*}$, while at 8B the contrast is unresolved ($+0.0233$, $\ci{-0.0246}{+0.0553}$) \cite{qwen3_2025,qwen3vl2025}. The advantage is not an artifact of the older baseline generation.

A development-bank $2\times2$ places the entire gain in the serialization layer (serializer $+0.059^{*}$/$+0.057^{*}$ under either perception; perception and interaction null), and three paired controls hold decoder and object facts fixed. First, the same VL weights reading serialized text instead of their own image gain $+0.0377^{*}$ at 7B and $+0.0949^{*}$ at 3B (confirmatory). Second, a decoder swap on identical text is null. Third, drawing every object's id, class, and range \emph{into the image} still leaves the VLM $0.124^{*}$ below those facts as text. An operator ladder attributes the block to side words and pre-banded ranges; a raw object list alone \emph{loses} $0.140^{*}$: decision-aligned text, not text per se, wins (tables in the supplement).

\subsection{What the Advantage Is Made Of}
\label{sec:mechanism}
Three preregistered probes (Amendment A3) decompose the gain.

\textbf{Not answer-string leakage.} On N1, with every stem and option reworded, the advantage persists: $+0.0414^{*}$ at 3B (attenuation $0.456$) and $+0.0738^{*}$ at 7B (larger than the confirmatory value; all 21 leave-one-sequence-out refits were significant). A wording-identity account predicts collapse; it did not occur.

\textbf{Dominated by boundary-aligned precomputation.} Replacing band strings with raw numeric ranges (D1) removes most of the advantage: the band vocabulary alone is worth $+0.0542^{*}$/$+0.0354^{*}$, 71\% and 83\% of the total. What survives numeric serialization is counting ($+0.1424^{*}$/$+0.1325^{*}$), the family whose options never shared vocabulary with the serializer. Coarse discretization at misaligned boundaries still helps a 3B reader (D2 vs D1, $+0.0158^{*}$): small readers benefit from any discretization, most from the aligned one.

\textbf{Bounded by the decision vocabulary.} On N2, whose judgment logic no serializer anticipated, the aligned serializer holds a small advantage at 7B ($+0.0413^{*}$) and none at 3B ($+0.0118$, unresolved); the numeric variant loses at 3B ($-0.0473^{*}$). Outside a known decision vocabulary, compilation cannot run ahead of the question, and small readers cannot do the arithmetic themselves. The interface is task-aware by construction, and this is its boundary.

\subsection{Reader Scale: Advantage, Floor, and Ceiling}
\label{sec:scale}

\begin{table}[t]
\centering
\caption{Reader-scale sweep (accuracy; cf = confirmatory bank). Pixel arms are cross-family, hence descriptive (A4). GT = ground-truth-state oracle. On N2, text arms cover 3971 of the 3974 items and pixel arms all 3974; contrasts are paired.}
\label{tab:scale}
\setlength{\tabcolsep}{4pt}
\footnotesize
\begin{tabular}{lcccc}
\toprule
reader & 0.5B & 1.5B & 3B & 7B \\
\midrule
\sys{} @ cf & 0.5143 & 0.7357 & 0.7673 & 0.7892 \\
GT @ cf & 0.6152 & 0.8660 & 0.9247 & 0.9699 \\
\sys{} @ N2 & 0.4132 & 0.4591 & 0.5014 & 0.5860 \\
D1 @ N2 & 0.4336 & 0.4666 & 0.4422 & 0.5963 \\
\midrule
& \multicolumn{4}{c}{\scriptsize moondream (1.4B) / llava-phi3 (3.8B) / Qwen2.5-VL (3B / 7B)} \\
pixel @ cf & \multicolumn{4}{c}{0.6114 \;/\; 0.6468 \;/\; 0.6913 \;/\; 0.7462} \\
pixel @ N2 & \multicolumn{4}{c}{0.2720 \;/\; 0.4867 \;/\; 0.4899 \;/\; 0.5448} \\
\bottomrule
\end{tabular}
\end{table}

\begin{figure}[t]
\centering
\includegraphics[width=0.95\linewidth]{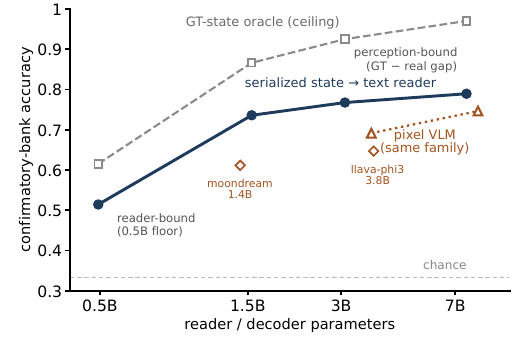}
\caption{Reader-scale sweep on the confirmatory bank. The serialized interface (solid) tracks its ground-truth-state ceiling (dashed) down to 1.5B and collapses at 0.5B, where even the oracle falls to $0.6152$: below 1.5B the regime is reader-bound; above 3B it is perception-bound (the gap to the ceiling, not reader capability, dominates). Pixel VLM arms (open markers) are shown at their own decoder scales; cross-family points are descriptive.}
\label{fig:scale}
\end{figure}

Table~\ref{tab:scale} lists accuracy at each reader size for the serialized interface (on the confirmatory bank and on N2), for the ground-truth oracle, and for the pixel VLMs; Fig.~\ref{fig:scale} plots the table's confirmatory-bank rows against model size. Down-scaling the reader from 7B to 1.5B costs little: at 1.5B the interface beats a comparable-size VLM ($+0.1242^{*}$, moondream \cite{moondream2024}) and a $2.5\times$ larger one ($+0.0889^{*}$, llava-phi3 \cite{liu2023llava,abdin2024phi3}); at 0.5B it collapses ($-0.0971^{*}$). The ground-truth oracle explains why: with perfect perception 0.5B reaches only $0.6152$ against $0.8660$ at 1.5B (step $-0.2508^{*}$), so the failure lies with reader capability, not with the interface. Above 3B the regime is \emph{perception-bound}; at 0.5B, \emph{reader-bound}; the crossover sits between 0.5B and 1.5B. On N2 every sub-7B arm struggles and the smallest VLM falls below chance ($0.2720$). A current-generation 2B thinking VLM under its native 1024-token budget (documented deviation) scores $0.7432$, descriptively.

\subsection{The Supervision Regime}
\label{sec:supervision}
Parity demands the VLM receive the supervision our detectors did: per fold, training QA comes from training-sequence frames via the frozen generator (no bank item enters training), and the VLM is LoRA-tuned \cite{hu2022lora} (QLoRA \cite{dettmers2023qlora} at 7B), hyperparameters from a fold-0 internal split, resolution control measured ($-0.0143^{*}$; runtime/quantization nil). We call the resulting arms FT-VLM, or FT-3B and FT-7B by size. Out-of-fold on the confirmatory bank they reach \textbf{0.8441 (FT-3B) and 0.8577 (FT-7B)}, overtaking the zero-shot serialized system at both sizes (serialized minus FT: $-0.0768^{*}$, $-0.0685^{*}$); the scale step is small ($+0.0136$, unresolved), so 16k samples nearly saturate at 3B.

Table~\ref{tab:ledger} shows the arms side by side with their training regimes. Supervision is not matched until the reader sees the same samples. A gradient-boosted option scorer \cite{chen2016xgboost} over the compiled state, fit on the development bank only, already reaches $0.8245$, beating the zero-shot 7B reader (reader $-$ scorer $=-0.0354^{*}$). \textbf{FT-Text}, the text reader LoRA-tuned on the same 16k items with the image replaced by the frame's serialized state and hyperparameters carried over, reaches $\mathbf{0.8396}$ out of fold: FT-Text $-$ FT-VLM $= -0.0045$ ($\ci{-0.0368}{+0.0177}$, no statistically resolved difference), while supervision buys the text side $+0.0723^{*}$ over its zero-shot value. Under matched supervision the choice of interface stops being an accuracy question and becomes one of compute placement, auditability, and zero-shot capability. The tuned VLM passes the serializer's own audits (N1 $+0.1258^{*}$, N2 $+0.0526^{*}$ over the zero-shot pixel arm): its gain is not format overfitting either.

Two findings survive supervision regardless. First, counting remains every route's weakest family (FT-VLM $0.586$/$0.606$; FT-Text $0.599$): the compiled state delivers that level with zero gradient samples ($0.566$/$0.583$), the VLM needs the full budget to draw level, and zero-shot the compiled counts win by $+0.1358^{*}$ at both sizes. Second, FT-7B ($0.8577$) remains $11.2$\,pp below the 7B ground-truth oracle ($0.9699$): both interfaces are ultimately perception-bound.

\begin{table}[t]
\centering
\caption{Supervision ledger (3B readers and VLMs; confirmatory bank). For every arm: how many gradient samples it was trained on, which input form it reads, and its accuracy. Full ledger in the supplement.}
\label{tab:ledger}
\setlength{\tabcolsep}{4pt}
\footnotesize
\begin{tabular}{lccc}
\toprule
arm & grad.\ samples & interface & acc \\
\midrule
\pixarm{} (zero-shot) & 0 & pixels & 0.6913 \\
\sys{} (zero-shot) & 0 & serialized & 0.7673 \\
gb-tree scorer & dev (shallow) & compiled state & 0.8245 \\
FT-VLM (LoRA, OOF) & 16k & pixels & 0.8441 \\
FT-Text (LoRA, OOF) & 16k & serialized & 0.8396 \\
\midrule
GT oracle (ceiling) & 0 & serialized (GT) & 0.9247 \\
\bottomrule
\end{tabular}
\end{table}

\subsection{What the Reader Buys}
\label{sec:readerbuys}
A deterministic option-blind answerer over the compiled state scores $0.619$; an option-aware learned scorer $0.8245$; the zero-shot 7B reader ($0.7892$) sits between, so the LLM is not the ceiling on the fixed families.

The learned scorers' advantage, however, does not survive the frozen probe banks. Table~\ref{tab:basefragile} provides the evaluation scores of the three option-aware scorers, unchanged, on the confirmatory bank and on both probe banks, with the zero-shot 7B reader as the reference row. Evaluated as-is on N1, with every stem and option reworded, all three option-aware baselines collapse to $0.3479$ overall, indistinguishable from chance ($1/3$): their features key on the exact option strings, and the paraphrase severs that link. On N2 they score $0.3183$ to $0.3510$, at chance again, with the gradient-boosted scorer (the strongest on the confirmatory bank) the weakest of the three. The zero-shot readers, by contrast, retain a resolved advantage on N1 ($+0.0414^{*}$/$+0.0738^{*}$, Section~\ref{sec:mechanism}) and the 7B reader reaches $0.5860$ on N2 (Table~\ref{tab:scale}). What the reader buys is therefore measured, not hypothesized: zero-supervision deployment at $0.7673$--$0.7892$ (3B--7B), whole-system robustness to reworded surface forms (not separately attributed), transfer to judgment vocabularies for which no scorer was trained, and aggregation under uncertainty, counting above all.

\begin{table}[t]
\centering
\caption{Option-aware learned baselines under the frozen probe banks (overall accuracy; fit on the development bank, evaluated as-is, no refit, no parser repair). Chance $=1/3$. cf = confirmatory bank.}
\label{tab:basefragile}
\setlength{\tabcolsep}{5pt}
\footnotesize
\begin{tabular}{lccc}
\toprule
scorer & cf & N1 & N2 \\
\midrule
rule & 0.8027 & 0.3479 & 0.3510 \\
logistic regression & 0.8080 & 0.3479 & 0.3332 \\
gradient-boosted trees & 0.8245 & 0.3479 & 0.3183 \\
\midrule
zero-shot 7B reader (\sys{}) & 0.7892 & 0.7914 & 0.5860 \\
\bottomrule
\end{tabular}
\end{table}

\subsection{Compute}
\label{sec:compute}
All compute numbers are measured, single stream, on one laptop GPU (per-module breakdown in Appendix~\ref{app:compute}). The perception stack totals ${\approx}0.63$B parameters, comparable to the baseline VLM's vision tower in parameter count, at ${\approx}0.95$\,s per frame. At the query stage the text reader answers in $0.151$--$0.241$\,s versus $1.109$\,s for the 7B VLM; at ten questions per frame, $2.79$--$3.27$\,s (3B--7B readers) versus $9.60$\,s re-encoding or $3.70$\,s with the VLM's visual prefix cached (the fair condition). Total compute is not smaller; it is placed \emph{outside} the reader, amortized across queries, and detachable to separate hardware. A reader-family swap (Llama-3.1-8B \cite{grattafiori2024llama3}) transfers at $-0.026^{*}$.

\section{Error Attribution: The Method Behind the Serializer}
\label{sec:attribution}

Because the serializer is deterministic, every question $q$ admits a \emph{closed-form answerer} $g(\cdot,q)$, the generator's own logic on any state, re-answerable from the perceived state with no LLM and no added annotation. With $y_q = g(\mathcal{S}^{\mathrm{GT}}, q)$ the gold answer, $\hat a_q$ the reader's answer, and $h_q = g(\hat{\mathcal{S}},q)$, each error splits into \emph{state-rule disagreement} ($\hat a_q{\neq}y_q \wedge h_q{\neq}y_q$), \emph{reader error} ($\hat a_q{\neq}y_q \wedge h_q{=}y_q$), and \emph{repairs} $C_q$ ($\hat a_q{=}y_q \wedge h_q{\neq}y_q$). The first term is \emph{state-rule disagreement}, not a causal perception error: $h_q$ being wrong shows only that the option-blind rule fails on the perceived state, and $C_q$ counts 310 items the reader repairs despite it; a paired ground-truth-versus-perceived counterfactual (supplement) gives the causal version.

Two findings drove the redesign. First, bearing calibration was not the primary source of error: 1 sector flip in 309; the losses were 43 undetected classes and 39 ordering errors, completeness rather than geometry. Second, most nearest-distance errors were \emph{reader} errors on correctly perceived states (54 of 98): the reader was handed $4.2$\,m and still picked the wrong band. That is the empirical basis of the serializer principle, confirmed by the operator ladder as a single-operator effect ($+0.258^{*}$).

\section{Discussion and Limitations}
\label{sec:discussion}

\textbf{Three regimes.} Zero-shot, the serialized interface wins, with the mechanism attributed (Section~\ref{sec:mechanism}) and the result replicated on a current model generation. Supervised, a tuned VLM overtakes the zero-shot system and an equally supervised text reader draws level, while compiled counting survives and the ledger (Table~\ref{tab:ledger}) keeps every comparison honest. At the ceiling, both routes are perception-bound: a 3B reader on ground-truth states ($0.9247$) exceeds a 7B reader on real ones ($0.7892$). Deleting suspect detections hurts ($-0.012^{*}$ at 7B) while an oracle deleting exactly the unmatched ones gains $+0.059^{*}$: precision is the largest lever, and the perceiver's confidence flag does not provide an effective mechanism for realizing it. The zero-shot margin narrows with VLM generation (4B $+0.0595^{*}$, 8B unresolved); auditability, a swappable perception stack, and a language-only reader do not depend on it.

\textbf{For the small-model use case.} The reader spends zero parameters on visual grounding, so a distilled student inherits language capability only; the scale study shows what that buys, and where it stops (0.5B). Under matched supervision the choice becomes engineering: zero-shot deployment, per-field auditability, amortized latency, and a swappable perception stack on one side; end-to-end simplicity and single-question latency on the other.

\textbf{Limitations.} (i) One domain; the confirmatory bank is temporally held out within the same recordings, not external validation. (ii) N2 quantifies the cost of leaving the known-vocabulary regime; open-ended queries remain future work. (iii) Nearest-class is a resolved zero-shot loss at 7B ($-0.0617^{*}$). (iv) Perception costs ${\approx}0.95$\,s per frame: query-amortized, not real-time. (v) The system uses stereo disparity while VLM arms get the monocular keyframe: a system-level, not sensor-parity, comparison. (vi) Supervised arms run at reduced resolution, cost measured ($-0.0143^{*}$). (vii) Contemporaneous continuous-control work \cite{zhou2026patchpolicy} (recent preprint) points the other way; its decisions are contact geometry with no symbolic vocabulary; ours are symbolic by construction. (viii) Cross-family pixel comparisons are descriptive only.

\section{Conclusion}
\label{sec:conclusion}

In this work, the visual tokens entering a language model are replaced with deterministic semantic serialization over a frozen perception stack, and the paper maps where the interface wins and where it does not. Zero-shot, it beats same-scale VLMs at both co-primary scales, survives paraphrase, replicates on a current model generation, and holds down to 1.5B readers; its advantage is decision-aligned precomputation, and its boundary is the decision vocabulary it can know in advance. With matched supervision the two interfaces converge ($0.8396$ vs $0.8441$, no resolved difference), counting stays the hardest family for every route, and every route remains perception-bound. For closed-set, object-level embodied QA, the practical statement is: vision can live outside the reader, the reader keeps every parameter for language, and between 1.5B and 7B, zero-shot, that trade wins outright.

\appendices
\section{Arm Glossary}
\label{app:arms}

Table~\ref{tab:arms} defines every evaluation arm referenced in the paper in one place. All arms decode greedily ($T{=}0$) on the forced three-choice track unless a deviation is noted; the supplement's amendment ledger records the full configuration of each run.

\begin{table}[t]
\centering
\caption{Arm definitions. cf = confirmatory bank; dev = development bank.}
\label{tab:arms}
\setlength{\tabcolsep}{4pt}
\scriptsize
\begin{tabular}{p{1.55cm}p{6.2cm}}
\toprule
arm & definition \\
\midrule
\pixarm{} 3B/7B & Qwen2.5-VL on the raw front-camera image plus the question \\
\pixarm{} @ N1/N2 & the same protocol run on the frozen N1/N2 probe banks \\
D0 (= \sys{}) & decision-scoped serializer: counts, corridor/overall arg-mins, per-class band and sector; per-object rows only for counting \\
D1 (numeric) & D0 with every distance as one-decimal meters, $\varepsilon$ flags dropped \\
D2 (numeric+alt) & D1 plus close/mid/far tags at 4/12\,m (disjoint vocabulary) \\
D3 (alt-only) & tags only, no numerals (negative control) \\
\devser{} & full per-object list serializer (pre-scoping production text) \\
GT-D0 & D0 rendered from ground-truth object states (perception removed) \\
BASE rule / logreg / gbtree & option-aware deterministic scorers over the compiled state, fit on the development bank only \\
blind & reader answers from question and options only, no scene \\
pixel sweep & moondream and llava-phi3 on cf and N2 (cross-family, descriptive) \\
pixel deviation & Qwen3-VL-2B under its native 1024-token thinking budget (descriptive only) \\
FT-VLM / FT-Text & LoRA-tuned VLM / text reader on the same 16k generator items, out-of-fold \\
\bottomrule
\end{tabular}
\end{table}

\section{Per-Family Contrasts and Blind-Arm Prior}
\label{app:perfam}

Table~\ref{tab:perfam} gives the complete per-family paired contrasts behind the summary in Section~\ref{sec:results}, under the same 21-cluster bootstrap ($B{=}20000$, fixed seed) as the primary endpoints. The two resolved gains, distance-band and counting, are the two families that reduce to deterministic arithmetic over the compiled state; nearest-class is a resolved loss at 7B; direction and path-object are unresolved at both scales.

\begin{table}[t]
\centering
\caption{Per-family paired accuracy differences, \sys{} $-$ \pixarm{}, confirmatory bank (cluster bootstrap, 95\% CIs). $^{*}$ = CI excludes zero.}
\label{tab:perfam}
\setlength{\tabcolsep}{4pt}
\scriptsize
\begin{tabular}{llccc}
\toprule
family & scale & $n$ & $\Delta$ & 95\% CI \\
\midrule
counting & 3B & 302 & $+0.1358^{*}$ & $\ci{+0.0413}{+0.2240}$ \\
counting & 7B & 302 & $+0.1358^{*}$ & $\ci{+0.0541}{+0.2022}$ \\
direction & 3B & 289 & $-0.0381$ & $\ci{-0.1308}{+0.0338}$ \\
direction & 7B & 289 & $-0.0242$ & $\ci{-0.0996}{+0.0393}$ \\
nearest-class & 3B & 243 & $+0.0412$ & $\ci{-0.0265}{+0.0940}$ \\
nearest-class & 7B & 243 & $-0.0617^{*}$ & $\ci{-0.1429}{-0.0089}$ \\
nearest-distance & 3B & 254 & $+0.2362^{*}$ & $\ci{+0.0821}{+0.3750}$ \\
nearest-distance & 7B & 254 & $+0.1890^{*}$ & $\ci{+0.1053}{+0.2766}$ \\
path-object & 3B & 240 & $+0.0042$ & $\ci{-0.0563}{+0.0530}$ \\
path-object & 7B & 240 & $-0.0417$ & $\ci{-0.1224}{+0.0126}$ \\
\midrule
all & 3B & 1328 & $+0.0761^{*}$ & $\ci{+0.0403}{+0.0996}$ \\
all & 7B & 1328 & $+0.0429^{*}$ & $\ci{+0.0073}{+0.0690}$ \\
\bottomrule
\end{tabular}
\end{table}

Table~\ref{tab:blind} shows the per-family breakdown of the blind-arm prior of Section~\ref{sec:audits}. The blind reader sees the question and options but no scene, so its accuracy bounds what surface cues alone can achieve after the audit gate; no family reaches the weakest scene-conditioned arm.

\begin{table}[t]
\centering
\caption{Blind-arm accuracy by family, confirmatory bank (chance $=1/3$).}
\label{tab:blind}
\setlength{\tabcolsep}{4pt}
\scriptsize
\begin{tabular}{lcccccc}
\toprule
arm & overall & counting & direction & near.-class & near.-dist & path-object \\
\midrule
blind 3B & 0.4142 & 0.374 & 0.439 & 0.342 & 0.492 & 0.425 \\
blind 7B & 0.4089 & 0.278 & 0.439 & 0.313 & 0.437 & 0.604 \\
\bottomrule
\end{tabular}
\end{table}

\section{A Serialized State, Verbatim}
\label{app:exemplar}

One real item, exactly as the reader saw it: the \sys{} serialization of a confirmatory-bank perceived state (question \texttt{coda\_0\_422\_nearest\_dist}), followed by its question. One exemplar per question family is included in the released package.

\begin{quote}
\footnotesize\linespread{0.92}\selectfont\ttfamily\raggedright
Counts by type: Bench=2 (2 uncertain), Bike=4, Bollard=7 (2 uncertain), Door=2 (2 uncertain), Floor Sign=2 (2 uncertain), Pole=2 (2 uncertain), Tree=2 (2 uncertain)\\
Closest to the robot's forward path: Bollard\\
Nearest object overall: Floor Sign (next nearest: Bollard)\\
Nearest of each type: Bench: over 15 meters away, in the center; Bike: over 15 meters away, on the left side; Bollard: 5 to 15 meters away, on the left side; Door: over 15 meters away, in the center; Floor Sign: 5 to 15 meters away, on the left side; Pole: over 15 meters away, in the center; Tree: over 15 meters away, on the left side
\end{quote}

\noindent Question: \emph{``How far is the nearest `Railing' from the robot?''} (gold: \emph{over 15 meters away}). Note what makes the item hard: the queried class, Railing, is \emph{absent} from the perceived state, a real detection miss. The reader must answer from a state that does not contain the object the question asks about; this is exactly the ``errors included'' regime the paper evaluates, and the counting rows, selections with runners-up, and pre-banded fields of Section~\ref{sec:serializer} are all visible above.

A second exemplar, \texttt{coda\_0\_436\_direction}, shows the opposite failure surface, a field that is present but disagrees with the gold:

\begin{quote}
\footnotesize\linespread{0.92}\selectfont\ttfamily\raggedright
Counts by type: Bench=2 (2 uncertain), Bike=4, Bollard=4, Chair=1 (1 uncertain), Door=2 (2 uncertain), Floor Sign=1 (1 uncertain), Pole=1 (1 uncertain), Railing=1, Scooter=1, Tree=1 (1 uncertain)\\
Closest to the robot's forward path: Floor Sign (next closest: Bollard)\\
Nearest object overall: Floor Sign (next nearest: Bollard)\\
Nearest of each type: Bench: over 15 meters away, on the right side; Bike: over 15 meters away, on the left side; Bollard: over 15 meters away, in the center; Chair: over 15 meters away, on the left side; Door: over 15 meters away, in the center; Floor Sign: 5 to 15 meters away, on the left side; Pole: over 15 meters away, on the left side; Railing: over 15 meters away, on the left side; Scooter: over 15 meters away, on the left side; Tree: over 15 meters away, on the left side
\end{quote}

\noindent Question: \emph{``Where is the nearest `Railing' located in the camera view?''} (gold: \emph{in the center}). Here the Railing field exists but reads ``on the left side'': an instance-selection and sector-rule disagreement of exactly the kind quantified in Section~\ref{sec:serializer} (question-level direction-field accuracy $0.553$ under the serializer's bearing rule).

Finally, a counting item, \texttt{coda\_0\_421\_counting}, shows the one stream that carries the per-object rows; every other family receives only the compiled header fields above (the scoping principle of Section~\ref{sec:serializer}):

\begin{quote}
\footnotesize\linespread{0.92}\selectfont\ttfamily\raggedright
Counts by type: Bench=2 (2 uncertain), Bike=4, Bollard=6 (2 uncertain), Chair=2 (2 uncertain), Cone=1 (1 uncertain), Door=1 (1 uncertain), Pole=2 (2 uncertain), Tree=2 (2 uncertain)\\
Closest to the robot's forward path: Bollard\\
Nearest object overall: Bollard (next nearest: Door)\\
Nearest of each type: Bench: over 15 meters away, in the center; Bike: over 15 meters away, on the left side; Bollard: 5 to 15 meters away, on the left side; Chair: over 15 meters away, in the center; Cone: over 15 meters away, in the center; Door: over 15 meters away, in the center; Pole: over 15 meters away, in the center; Tree: over 15 meters away, on the left side\\
Visible objects (nearest first):\\
\#1 [Bollard] dist=7.81m bearing=36.2deg (left) view=left (uncertain)\\
\#2 [Door] dist=16.97m bearing=-11.2deg (center) view=center (uncertain)\\
\#3 [Pole] dist=17.71m bearing=-8.4deg (center) view=center (uncertain)\\
\#4 [Cone] dist=17.85m bearing=-8.3deg (center) view=center (uncertain)\\
\#5 [Bollard] dist=17.91m bearing=-35.8deg (right) view=right (uncertain)\\
\#6 [Bench] dist=18.2m bearing=-7.7deg (center) view=center (uncertain)\\
\#7 [Tree] dist=18.56m bearing=36.2deg (left) view=left (uncertain)\\
\#8 [Bollard] dist=18.75m bearing=4.8deg (center) view=center\\
\#9 [Bollard] dist=19.26m bearing=25.1deg (left) view=left\\
\#10 [Chair] dist=19.68m bearing=6.9deg (center) view=center (uncertain)\\
\#11 [Bike] dist=21.02m bearing=26.3deg (left) view=left\\
\#12 [Bollard] dist=21.1m bearing=30.1deg (left) view=left\\
\#13 [Bike] dist=21.21m bearing=24.3deg (left) view=left\\
\#14 [Pole] dist=21.21m bearing=24.4deg (left) view=left (uncertain)\\
\#15 [Bike] dist=21.53m bearing=17.3deg (left) view=left\\
\#16 [Bench] dist=21.78m bearing=-16.3deg (right) view=right (uncertain)\\
\#17 [Bollard] dist=22.22m bearing=34.3deg (left) view=left\\
\#18 [Chair] dist=23.59m bearing=25.3deg (left) view=left (uncertain)\\
\#19 [Bike] dist=27.31m bearing=35.6deg (left) view=left\\
\#20 [Tree] dist=30.7m bearing=36.2deg (left) view=left (uncertain)
\end{quote}

\noindent Question: \emph{``How many objects of type `Trash Can' are visible in the camera view?''} (gold: \emph{1}). Once more, the queried class is entirely absent from the perceived state, in both the count header and the rows: the reader must answer a count question about an object the perceiver never reported. Items like this are why counting remains the weakest family for every route in Section~\ref{sec:supervision}. The exemplars were not curated for success; they are the first confirmatory items from their respective families in the released package.
\section{Per-Module Compute}
\label{app:compute}

Table~\ref{tab:modules} shows the parameter count and measured single-stream latency of every perception-stack module (RTX 5090 Laptop GPU, 200 confirmatory frames each). The stereo block runs on the CPU and can overlap the GPU modules; the serial sum gives the conservative ${\approx}0.95$\,s total of Section~\ref{sec:compute}. Peak per-module GPU memory is $4.2$\,GiB. Table~\ref{tab:latency} gives the measured per-question latency of the readers and VLMs (greedy decoding, 30 questions after warm-up).

\begin{table}[t]
\centering
\caption{Perception-stack modules: parameters and measured mean latency per frame (single stream; 200 frames).}
\label{tab:modules}
\setlength{\tabcolsep}{5pt}
\scriptsize
\begin{tabular}{lcc}
\toprule
module & parameters & mean latency \\
\midrule
YOLO11l + YOLO11m, full frame & 25.3\,M + 20.1\,M & 62.5\,ms \\
YOLO11l + YOLO11m, $2{\times}2$ tiling & (same weights) & 221.3\,ms \\
LLMDet (absence recovery only) & 233.0\,M & 298.0\,ms \\
UniDepthV2 (monocular depth) & 353.8\,M & 98.7\,ms \\
stereo semi-global block matching (SGBM, CPU) & 0 & 267.7\,ms \\
semantic serializer (code) & ${\approx}0$ & 0.1\,ms \\
\midrule
total & ${\approx}0.63$\,B & ${\approx}0.95$\,s (serial) \\
\bottomrule
\end{tabular}
\end{table}

\begin{table}[t]
\centering
\caption{Per-question latency (mean / median, s).}
\label{tab:latency}
\setlength{\tabcolsep}{5pt}
\scriptsize
\begin{tabular}{lcc}
\toprule
model & mean & median \\
\midrule
Qwen2.5 0.5B / 1.5B (text) & 0.151 / 0.181 & 0.151 / 0.175 \\
Qwen2.5 3B / 7B (text) & 0.196 / 0.241 & 0.195 / 0.230 \\
Qwen2.5-VL 3B / 7B (VLM) & 0.972 / 1.109 & 0.970 / 1.102 \\
\bottomrule
\end{tabular}
\end{table}

\section{Bank Composition and Separation Audit}
\label{app:bank}

Table~\ref{tab:bank} summarizes the composition of the confirmatory bank: the five question families are nearly balanced, and the 21 recording sequences that serve as bootstrap clusters range from 1 to 254 questions (median 43; the largest, sequence 20, motivates the cluster-level resampling of \eqref{eq:boot} and the leave-one-sequence-out refits of Section~\ref{sec:mechanism}). Every confirmatory frame is at least $1.5$\,s from every development frame (median $2.1$\,s, P95 $4.7$\,s, max $11.4$\,s; $58.6\%$ at least $2$\,s, $4.6\%$ at least $5$\,s).

\begin{table}[t]
\centering
\caption{Confirmatory-bank composition (1328 questions, one per frame).}
\label{tab:bank}
\setlength{\tabcolsep}{5pt}
\scriptsize
\begin{tabular}{lc}
\toprule
quantity & value \\
\midrule
counting / direction / nearest-class & 302 / 289 / 243 \\
nearest-distance / path-object & 254 / 240 \\
gold letters A / B / C / D (4-option bank) & 337 / 356 / 316 / 319 \\
cluster sizes (21 sequences), min / median / max & 1 / 43 / 254 \\
separation from dev bank, min / median / P95 / max & 1.5 / 2.1 / 4.7 / 11.4\,s \\
\bottomrule
\end{tabular}
\end{table}

\section*{Acknowledgment}
The authors used a large language model (Claude, Anthropic) as a writing and editing assistant during the preparation of this manuscript. All experiments, data, analyses, and claims were designed, executed, and verified by the authors, who take full responsibility for the content.

\section*{Data and Code Availability}
The question banks, per-item prediction logs for every arm, perceived-state files, serialized-state exemplars, the frozen preregistration amendments, and all analysis scripts are released at \url{https://github.com/drxucong/semantic-serialization-scene-qa} (data CC BY-NC-SA 4.0, inherited from CODa; code MIT); every statistic reported in this paper can be recomputed from the released per-item logs. Throughout, \emph{the supplement} denotes \texttt{SUPPLEMENT.md} in that release, which locates the material behind each pointer in the text. The perception stack and serializer are specified to reproduction detail in Section~\ref{sec:system} and the supplement; trained detector weights are available on reasonable request. CODa images are not redistributed.

\bibliographystyle{IEEEtran}
\bibliography{refs}

\nobreak\vskip -3.5\baselineskip plus -1fil\relax
\begin{IEEEbiography}[{\includegraphics[width=1in,height=1.25in,clip,keepaspectratio]{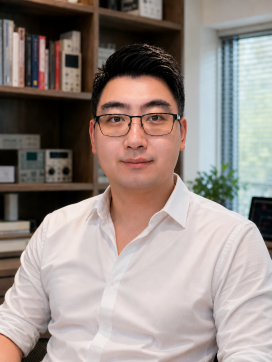}}]{CONG XU}
\looseness-1 received the B.S. degree in electrical engineering and automation from Zhengzhou University, Zhengzhou, China, and the M.S. degree in electrical engineering from the University of South Florida, Tampa, FL, USA, in 2020, where he is currently pursuing the Ph.D. degree in electrical engineering with the Department of Electrical and Computer Engineering, as a member of the Interdisciplinary Communications, Networking and Signal Processing (iCONS) Lab. His M.S. thesis, supervised by Prof.\ Ravi Sankar, developed a spatial stereo acoustic source-localization system with an optimized three-dimensional time-difference-of-arrival sensor arrangement and convolutional-neural-network-based source recognition over spectrograms, motivated by the problem of perceiving moving objects that are occluded or outside the line of sight, an early step toward the multimodal perception theme of his doctoral work.

\looseness-1 His doctoral research asks how language-model reasoning can be brought onto resource-constrained embodied platforms, and spans the full stack from perception algorithms and language-model interfaces to hardware implementations. His research interests include language--vision collaborative SLAM, structured world models for embodied agents, cross-modal risk perception under sensor degradation, deterministic semantic serialization as a perception interface for small language models, and efficient on-device AI on FPGA and SoC platforms, including the power--performance trade-offs of AI accelerators and edge computing systems, with an emphasis on prospectively frozen protocols and reproducible, audit-friendly evaluation.

\looseness-1 Professionally, he serves as a Technology Specialist at Global Electronics Testing Services (Global ETS), where he leads work on AI-driven silicon validation and custom ASIC design; his recent project, TurboMemory-X, targets memory bandwidth and cost efficiency for AI accelerators as an alternative to high-cost high-bandwidth-memory solutions. He is the Founder and Chairman of the International Association of Hybrid Artificial Intelligence (IAHAI), a nonprofit organization. He is a regular speaker at SMTA venues on semiconductor supply-chain security and hardware reliability: at the SMTA Symposium on Counterfeit Parts and Materials, he and his co-researchers presented an AI-powered hardware-testing methodology for counterfeit-component detection, and he presents on DRAM reliability, hardware telemetry, and screening protocols for next-generation AI systems. He also serves as a Guest Editor of an MDPI Special Issue on Intelligent Transportation Systems and Its Applications.
\end{IEEEbiography}

\makeatletter
\let\origIEEEneedspace\@IEEEtranneedspace
\def\@IEEEtranneedspace#1#2{\relax}
\makeatother
\nobreak\vskip -2.5\baselineskip plus -1fil\relax
\begin{IEEEbiography}[{\includegraphics[width=1in,height=1.25in,clip,keepaspectratio]{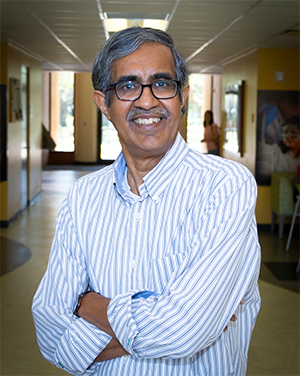}}]{RAVI SANKAR}
\looseness-1 (Life Senior Member, IEEE) received the B.E. (Honors) degree in electronics and communication engineering from the University of Madras, India, the M.Eng. degree in electrical engineering from Concordia University, Canada, and the Ph.D. degree in electrical engineering from The Pennsylvania State University, USA. He has been with the Department of Electrical and Computer Engineering, University of South Florida, Tampa, FL, since 1985, where he is currently a USF Theodore and Venette Askounes-Ashford Distinguished Scholar award-winning Professor and the Director of the interdisciplinary Communications, Networking and Signal Processing (iCONS) research lab. He was a 2015--16 Fulbright Fellow to Brazil to conduct collaborative research focusing particularly on the use of multiple sensors and signal processing to improve healthcare, and was a visiting research fellow of the Japanese Society for the Promotion of Science (JSPS), nominated by the NSF in spring 2000 to conduct collaborative research in Japan. He also held visiting positions at the University of Melbourne, Australia, in summer 2000, the U.S. Air Force Research Lab (Rome Lab), Rome, NY, in summer 1997, and Motorola, Boynton Beach, FL, in summer 1991.

\looseness-1 Prof.\ Sankar's main research interests are in the areas of wireless communications, networking, signal processing, and their applications. His current focus is on advancing healthcare and intelligent systems through multimodal sensing, signal processing, machine learning and AI, wearable technologies, and edge intelligence. He has published extensively in those areas with over 250 papers in journals and premier international conferences and several book chapters. His research has been cited widely, as measured by an h-index of 33 and an i10-index of 93. Through the years, he has supervised 8 post-doctoral researchers, over 72 Ph.D. and M.S. students, and numerous B.S. senior capstone design projects. Further, he has mentored many more students, including another 15 Ph.D. students, and is currently directing 3 Ph.D. students. The iCONS research lab under his leadership has successfully conducted numerous funded research projects over the years with support from various federal and state agencies and industries. Research contributions of the lab have been widely recognized for their quality and reputation and further advanced by fostering major international collaborations in South Korea and Brazil.

\looseness-1 Prof.\ Sankar was a Distinguished Lecturer for the IEEE Engineering in Medicine and Biology Society (EMBS) in 2014--16. He has delivered numerous (more than 50) plenary, keynote, or invited lectures at international conferences and institutions over the years all over the world, including Korea, Japan, Mexico, Brazil, and India. He has received numerous awards, including the IEEE Florida Council Outstanding Engineering Educator award and the Outstanding Contributions in Research award from the ASEE. He has served on the editorial board of several journals, including as an Associate Editor of \emph{IEEE Communications Surveys and Tutorials}, on organizing committees and technical program committees, and as a session organizer and chair for many flagship IEEE conferences. He was the organizer and co-chair of the US--Korea Joint International Workshop on Global Wireless Sensor Networks (GWSN), sponsored by the NSF and KOSEF (Korea Science and Engineering Foundation), held in Korea in 2009 and 2011. He has served the IEEE in various capacities, such as the Founding Chair of the Engineering in Medicine and Biology Society (EMBS) chapter of the IEEE Florida West Coast Section and the Vice-Chair of the IEEE Signal Processing Society chapter.
\end{IEEEbiography}
\makeatletter\let\@IEEEtranneedspace\origIEEEneedspace\makeatother

\EOD

\end{document}